\documentclass[runningheads]{llncs}
\usepackage{graphicx}
\usepackage{subfig}
\usepackage{amsmath}

\usepackage{hyperref}
\hypersetup{colorlinks=true, allcolors=blue}

\begin{document}

\title{Markerless Augmented Advertising for Sports Videos\thanks{Supported by the Institute for Pure and Applied Mathematics (IPAM) at the University of California, Los Angeles, GumGum Inc. and U.S. National Science Foundation Grant DMS-0931852.}} 
\titlerunning{Markerless Sports Advertising} 


\author{
	Hallee E. Wong\inst{1} 
	\and
	Osman Akar\inst{2} 
	\and
	Emmanuel Antonio Cuevas\inst{3}
	\and 
	Iuliana Tabian\inst{4}
	\and
	Divyaa Ravichandran\inst{5}
	\and
	Iris Fu\inst{5}
	\and
	Cambron Carter\inst{5}
}


\authorrunning{H. E. Wong et al.} 


\institute{Williams College, Williamstown, USA \email{hew1@williams.edu} \\
\and
University of California, Los Angeles, USA \email{osmanakar1123@hotmail.com} \\ 
\and
Universidad de Guanaquatro, Guanajuato, Mexico
\email{emmanuel.antonio@cimat.mx}\\ 
\and
Imperial College London, UK \email{iuliana.tabian15@imperial.ac.uk}\\
\and
GumGum Inc., Santa Monica, USA \email{\{dravi,iris,cambron\}@gumgum.com}
}

\maketitle


\begin{abstract}
Markerless augmented reality can be a challenging computer vision task, especially in live broadcast settings and in the absence of information related to the video capture such as the intrinsic camera parameters. This typically requires the assistance of a skilled artist, along with the use of advanced video editing tools in a post-production environment. We present an automated video augmentation pipeline that identifies textures of interest and overlays an advertisement onto these regions. We constrain the advertisement to be placed in a way that is aesthetic and natural. The aim is to augment the scene such that there is no longer a need for commercial breaks. In order to achieve seamless integration of the advertisement with the original video we build a 3D representation of the scene, place the advertisement in 3D, and then project it back onto the image plane. After successful placement in a single frame, we use homography-based, shape-preserving tracking such that the advertisement appears perspective correct for the duration of a video clip. The tracker is designed to handle smooth camera motion and shot boundaries.

\keywords{Augmented reality \and Virtual advertisement insertion \and Overlay video advertising}
\end{abstract}


\section{Introduction}

Advertisements traditionally occur in between breaks during the broadcast of a live sporting event. These breaks interfere with the viewing experience enough to motivate the advent of Digital Video Recording (DVR). This technology allows viewers to skip through commercial breaks, which negates the impact of the advertising altogether. This work motivates a system which integrates advertising seamlessly into a live sports broadcast without diverting too much attention from the main focus in the stream\textemdash namely the action taking place on the field, court, rink, etc.

Different approaches have been attempted to blend an advertisement, which we will refer to as \textquotedblleft asset\textquotedblright, into video content. The goal is to place an asset such that augmentation has a minimal impact on the viewing experience. Manual solutions exist but require the expertise of a graphic artist. While effective, this approach is laborious and expensive. Alternative methods such as visual attention \cite{Chang:2008:VVS:1459359.1459500} and visual harmony \cite{Liu2009AdvertiseG} have been used to reduce human dependency while achieving similar computational aesthetics. In contrast to common video asset overlays that consistently reside in the same position, often resulting in occluding important content information, the work by Mei et al. \cite{Mei:2010:ATC:2736420.2736646} detects non-salient regions in the video before overlaying relevant assets based on textual descriptions. Similarly, Chang et al. \cite{Chang2010VirtualSA} devise an algorithm to identify regions with high viewer interest and insert assets exhibiting color harmony with the content to stay visually consistent. While both of these methods strive to minimize the lack of coherence between the asset placement and the video content, the final asset can be argued to take away from the focus of the content despite the intended purpose \cite{Mei:2008:CIA:1459359.1459418}. In the examples provided by \cite{Chang2010VirtualSA} and \cite{10.1007/978-3-540-73417-8_52}, the asset placement is on the tennis court itself because the system heavily relies on the court lines for obtaining camera information. However, this asset placement can interfere with the focus of the video - which is the tennis ball in the game. 

Placing assets in and nearby the focus of the video can compromise the viewing experience. We take an alternative approach: we leverage underutilized virtual real estate which we can automatically detect. In this work, we identify ``virtual real estate" as the crowded sections of a sporting arena and use those regions for asset insertion. We argue that the latter are good candidates for asset placement because they are of less relevance to the focus of the broadcast \cite{Li:2005:RTA:1101149.1101221,Wan2006AutomaticCP}.

In this work, we propose a novel system that: 
\begin{enumerate}
	\item Automatically identifies viable ``crowd"
regions through spatio-temporal analysis in a video clip 
	\item Places the asset in a ``natural"
fashion, respecting the physical constraints of the real world scene 
	\item Is fully automatic 
\end{enumerate}

The rest of this work is organized as follows: section \ref{sec:Related-Work} discusses some relevant work on the topic, section \ref{sec:Setup_assumptions} discusses our system and each of its parts in detail, and we conclude with the results and discussions in section \ref{sec:Conclusion}. 

\section{\label{sec:Related-Work}Related Work}

Various systems and pipelines have been proposed for the insertion of assets in video. Ideally, the asset should integrate seamlessly without occluding any pertinent video content. The asset should be apparent but not disruptive. ImageSence \cite{Mei:2008:CIA:1459359.1459418} solves this problem by choosing insertion content according to visual consistency. The authors in \cite{Chang:2008:VVS:1459359.1459500} work around the problem by harmonizing the asset to be visually consistent with the original content. Liu et al. in \cite{Liu2009AdvertiseG} define intrusiveness as follows: a) if the inserted asset covers the Region of Interest (ROI), it is truly very intrusive, and b) if the asset distracts audience attention from the original attending point, it is also very intrusive. They use these metrics to insert assets ``gently" into the target content.

One of the challenges in achieving augmented advertising is constraining the overlaid asset to the physical boundaries of the real world scene. Xu et al. \cite{10.1007/978-3-540-30542-2_33} rely on static regions (e.g. ground) in a soccer field and make use of strong and reliable field boundary markings so that the final asset could be placed appropriately. Wan et al. \cite{Wan2006AutomaticCP} use the geometry of the candidate regions derived from known markers to decide whether a text, image or animation type of asset is more suitable. Chang et al. \cite{Chang2010VirtualSA} represent the asset in the chosen video frame by color-harmonizing it with the background. Given the known landmarks, and calibration parameters, derived 3D camera pose is used to project the asset onto the region of interest. They also use a visual attention model to find the most ``attention-grabbing'' region to place the asset. However, it can make the final asset placement very intrusive to the viewing experience, since in their specific use case of tennis matches, the highest attention region is around the player. 

Our biggest challenge was to achieve the goal of virtual asset placement in compliance with the conditions of a) non-intrusiveness, and b) conformity to real world scene constraints, while requiring minimal manual intervention. Usually, advertisers hire professional editors to manually implement virtual ads. It usually is very labor-intensive and inefficient for rapid productions on monetizing sports videos. We approach the problem with the intention of having a fully automatic pipeline. The authors in \cite{Wan2006AutomaticCP} aim to automate the pipeline for sports highlights; however, there is an initial manual step of segmenting out the highlights in the video. They also model their virtual reality around the boundaries of a single game event, so they manually remove the replays from the broadcast videos as well. The work done in \cite{10.1007/978-1-4471-0859-7_6} is similar in attempting to construct a system with a fundamental requirement being that the system performs on-line in real-time, with no human intervention and no on screen errors. Prior assumption that the billboard to be substituted in the target video is known beforehand reduces the problem to one of template matching. 

It must be noted that a lot of the work briefly reviewed in this section are built around certain very specific use cases (tennis \cite{Chang2010VirtualSA,Chang:2008:VVS:1459359.1459500}, soccer \cite{10.1007/978-3-540-30542-2_33}, baseball \cite{Li:2005:RTA:1101149.1101221}), relying heavily on the presence of known, reliable markings on the ground and a rigid sporting-ground structure enabling assets to be integrated seamlessly into the scenery. 

We aim to be independent of such assumptions while focusing on the crowds and the surrounding areas at sporting events. Our approach also completely removes the manual component from the entire system, presenting an end-to-end pipeline for the overlay of ads in a non-intrusive, engaging fashion.

\section{\label{sec:Setup_assumptions}Setup}

The assumptions that are made for the proposed system are detailed in the following list:
\begin{itemize}
	\item the input video is captured from a single viewpoint with a monocular RGB camera
	\item intrinsic parameters about the camera may not be known 
	\item the texture we aim to overlay on will be a crowd of people in a sports stadium 
	\item the asset to be overlayed in the video will be a 2D image/video and not a 3D object 
	\item the shot need not be static, i.e., the camera may change location or angle 
	\item the video may contain multiple camera shots 
\end{itemize}

\begin{figure}
	\centering
	\includegraphics[width=0.7\linewidth]{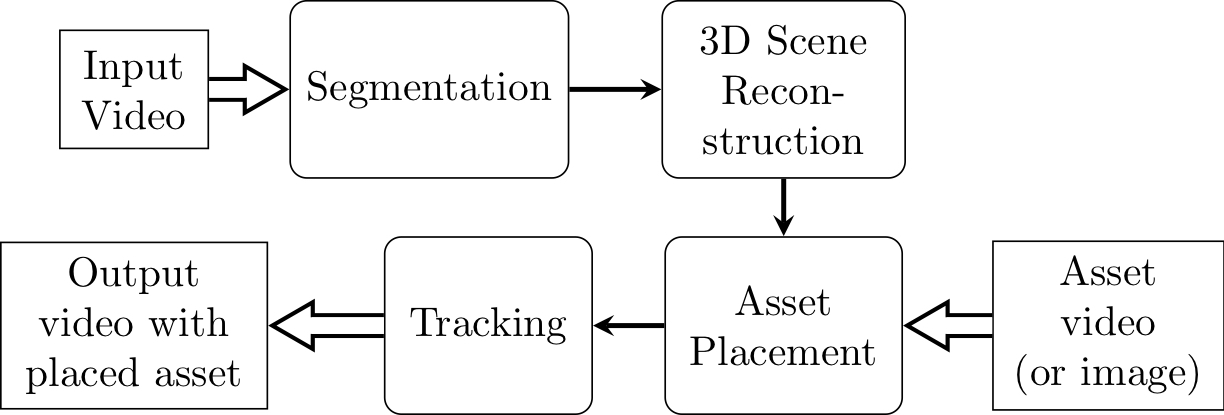}
	\caption{Our proposed automated pipeline for augmentation.}
	\label{fig:System-block-diagram}
\end{figure}

The following sections describe the system in detail and the above assumptions are addressed in more detail. Fig. \ref{fig:System-block-diagram} illustrates a conceptual diagram of our proposed pipeline.

\subsection{Segmentation and Seed Frame Selection}

To augment the input video efficiently, we first attempt to isolate and overlay augmentation on the most ideal frame. Once this frame is determined, we track our asset through subsequent frames. We are primarily concerned with whether the seed frame contains the texture of interest (e.g. crowds in sport stadium imagery) and if the textured area is large enough for the asset to be adequately overlayed.

Semantic segmentation decomposes a scene into its various components at the pixel level. Our system requires pixel-level  in order to adequately overlay an asset onto the scene. This work makes use of a convolutional neural network (CNN) based segmentation technique called Pyramid Scene Parsing network (PSPNet) \cite{zhao2016pyramid}, which has been shown to be competitive in various scene-parsing challenges (e.g. ImageNet 2016 \cite{ILSVRC15}, PASCAL VOC 2012 \cite{pascal-voc-2012}, and CityScapes 2016 \cite{Cordts2016Cityscapes} benchmarks). 

\begin{figure}
	\centering
	\subfloat[Input image]
	{
		\frame{\includegraphics[width=0.35\textwidth]{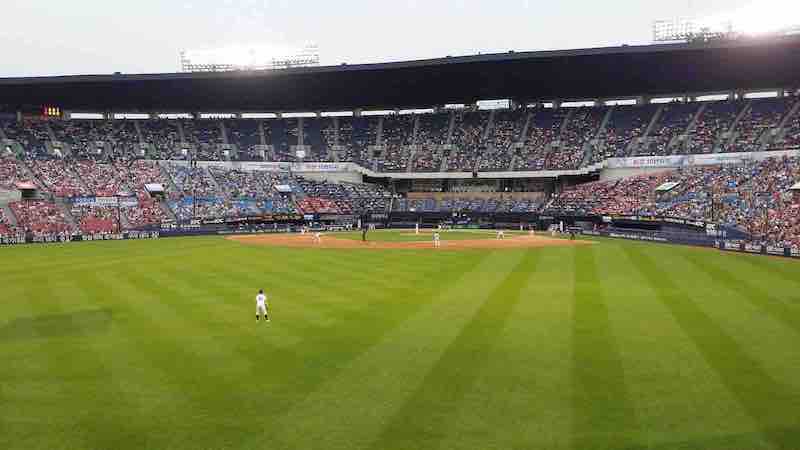}}
		\label{fig:action}
	} 	
	~
	\subfloat[Image segmented by PSPNet]
	{
		\frame{\includegraphics[width=0.35\textwidth]{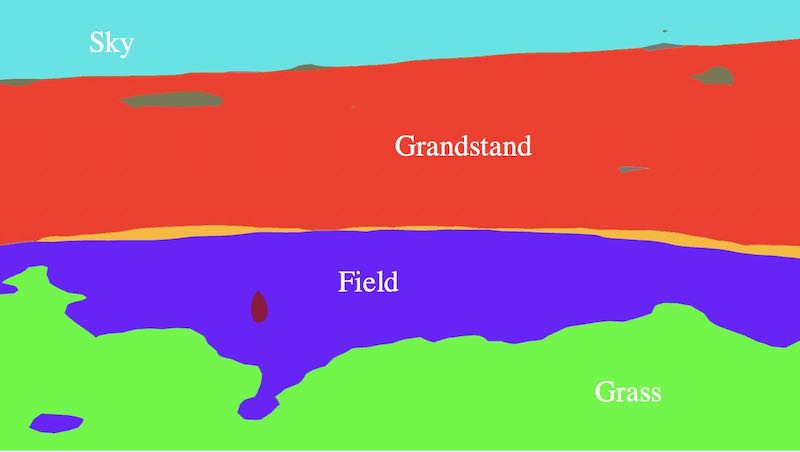}}
	}
	\caption{The input image of a baseball game is segmented by PSPNet's ADE20K trained model into sky (\emph{blue}), grandstand (\emph{red}), wall (\emph{gray}) fence (\emph{orange}), field (\emph{purple}), and grass (\emph{green}) \cite{zhou2016semantic,zhou2017scene}.}
	\label{fig:PSPseg}
\end{figure}

We make use of a PSPNet model trained on the ADE20K \cite{zhou2016semantic,zhou2017scene} dataset subset to identify our textures of choice: ``person" and ``grandstand". We refer to this combination of classes as ``crowd". Fig. \ref{fig:PSPseg} shows an example segmentation using PSPNet.

This segmentation technique is computationally expensive to carry out on every frame. To reduce the computational overhead, we sample the video at one frame per second and only segment those frames. Alternative segmentation techniques could be explored to alleviate computation time but we will restrict our focus to accuracy using this method. Once we have identified a seed frame, we discontinue segmentation until tracking fails, which will be discussed in Sec. \ref{subsec:Asset-Tracking-Across}. 

\subsubsection{Measuring Segmentation Quality}\label{subsec:Measuring-segmentation-quality}

Due to errors in judgment in classifying the image at the pixel level (segmentation), the segmentation mask can be far from ideal, containing holes, fragmentation etc. We aim to ensure that a) the asset only covers the crowd region of the image and b) the crowd region of the image is large enough and contiguous enough such that further processing has a higher likelihood of succeeding. Thus we define an ideally segmented region to be one which:
\begin{itemize}
	\item Is connected 
	\item Contains no holes 
	\item Is compact
\end{itemize}

To evaluate the binary segmentation images of the sampled frames with respect to these three conditions, we propose a metric called the Segmentation Quality Score (SQS). The SQS is defined as the product of three subscores,
\begin{equation}
	SQS = S_{cp} \ S_{cl} \ S_{sp} 
	\label{eq:SQS}
\end{equation}
where $S_{cp}$ is the component score, $S_{sp}$ is the completeness score and $S_{sp}$ is the shape score. Low values are associated with good segmentation and large values are associated with poor segmentation. The minimum possible SQS score is 1. 

\paragraph{Component Score}

To evaluate the connectedness of the segmented crowd region, we calculate the component score $S_{cp}$ as 

\begin{equation}
	S_{cp} \ = \ \frac{ a }{ \max_{i} a_i } \ ,
	\label{eq:component}
\end{equation}
where $a$ is the area of the entire crowd region and $a_i$ is the area of the $i^{th}$ connected component in the crowd region. The image is divided into disjoint connected components and the largest component is identified. Next, the ratio of the area of this largest component with respect to the entire crowd region in the binary image is computed. This value will always be larger than or equal to 1 (which will be the case when there is only one major component). 

\paragraph{Completeness Score}

The completeness score $S_{cl}$, defined as 

\begin{equation}
	S_{cl} \ = \ \frac{a'}{a}
	\label{eq:completeness}
\end{equation}
where $a'$ is the total area of the crowd region with all holes filled in, quantifies the size of hole(s) in the segmented area. This score is close to 1 when the total area of holes in the larges crowd segmented crowd component is small. 

\paragraph{Shape Score}

\begin{figure}
	\centering
	\subfloat[${SQS} = 1.85$]
	{
		\frame{\includegraphics[width=0.25\linewidth]{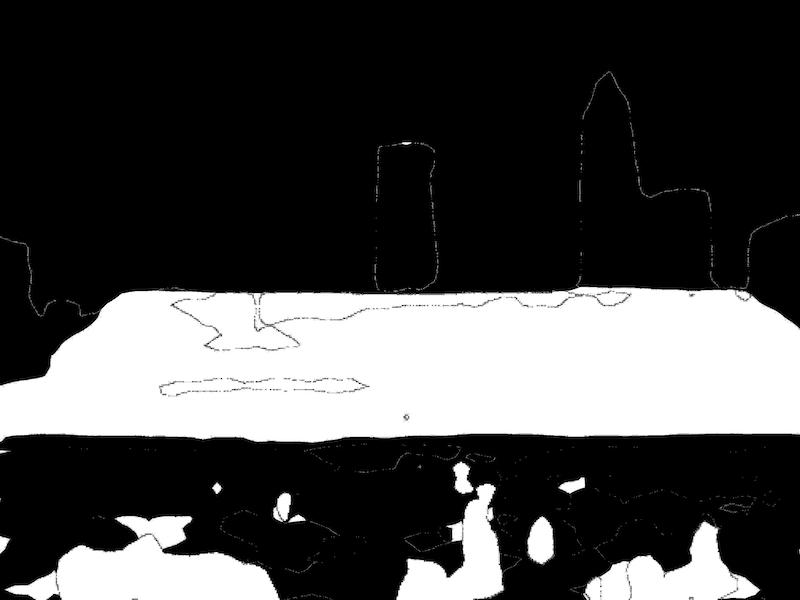}}
	}
	~
	\subfloat[${SQS} = 2.58$]{
		\frame{\includegraphics[width=0.25\linewidth]{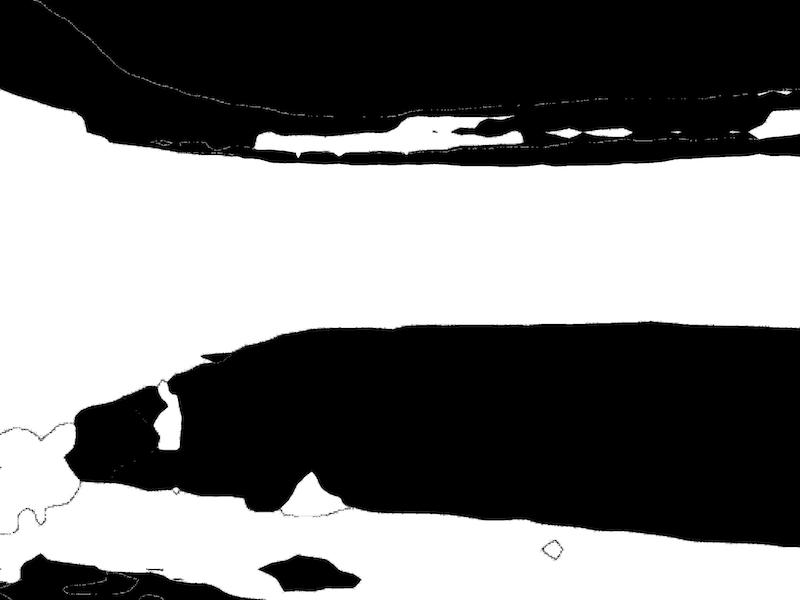}}
	}
	~
	\subfloat[${SQS} = 4.43$]{
		\frame{\includegraphics[width=0.25\linewidth]{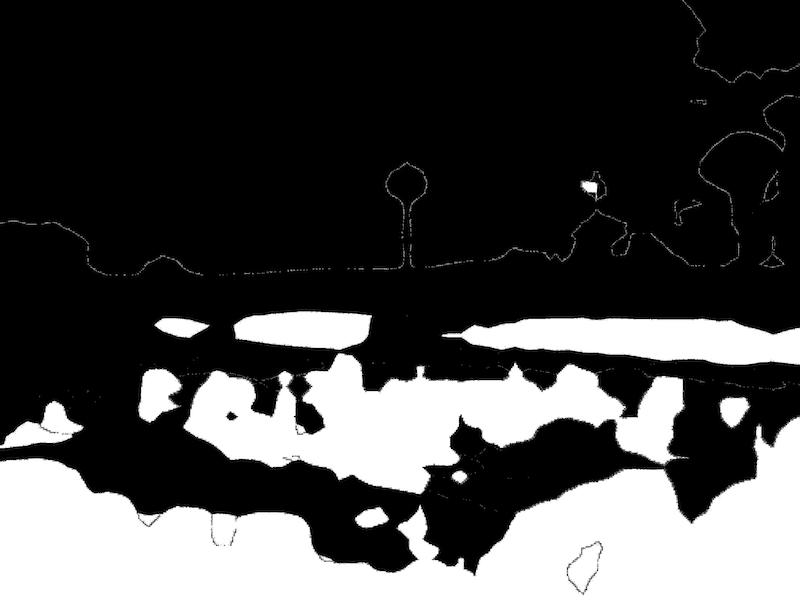}}
	}
	\caption{Segmented images and the SQS associated with the quality of the segmentation. The smaller the SQS value, the better the quality}
	\label{fig:shape_score}
\end{figure}

The shape score, defined as 
\begin{equation}
	S_{sp} \ = \ \frac{ p_j }{ 2 \ \sqrt{ \pi \ a_j}}
	\label{eq:shape}
\end{equation}
where $p_j$ and $a_j$ are the perimeter and area of the largest connected crowd component respectively, quantifies the regularity of the shape of the largest connected crowd component. If the crowd region has a large boundary relative to the area it covers, it is indicative of an irregular shape (Fig. \ref{fig:shape_score}). The denominator (Eqn. \ref{eq:shape}) contains the $\sqrt{ \ a_j}$ term to make the score invariant under scaling and a factor of $2\sqrt{\pi}$ so that the minimum value of $S_{sp}$ is 1 (when the shape is circular). We assume that the crowd will be seated or standing in a regularly shaped area in the images, so if the crowd segmentation is highly irregular, we conclude that the segmentation is not suitable for an asset placement and likely to be erroneous. 

For some frames with small SQS, the detected crowd area is significantly small. To eliminate the chance of trying to further process such cases, the algorithm first runs through all selected frames and finds the maximum detected crowd area, then disregards all frames with detected crowd area less than the half of the maximum area. The frame with the lowest SQS (Eqn. \ref{eq:SQS}) is then chosen from the remaining frames as the best-suited frame for asset placement.


\subsection{3D Reconstruction}

After deciding on a keyframe for asset placement, we estimate the depth of locations in the scene relative to the camera. This will facilitate in identifying the dominant plane coinciding with the crowd region and placing the asset on this plane in a perspective-correct way.

\subsubsection{Depth Estimation}

\begin{figure}
	\centering
	\subfloat[Input image]
	{
		\frame{\includegraphics[width=0.35\textwidth]{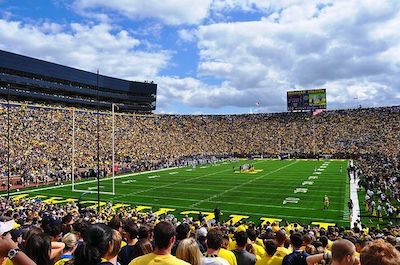}}
	} 
	~
	\subfloat[Inverse depth map]
	{
		\frame{\includegraphics[width=0.35\textwidth]{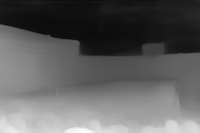}}
	}
	\caption{An example of a crowd image and a inverse depth map visualization of the relative depth values predicted by MegaDepth. In (b) dark pixels have large predicted relative depths and light pixels have small relative predicted depth values.}
	\label{fig:An-example-ofMegadepth}
\end{figure}

To create a 3D reconstruction of the scene, we must first estimate the distance between the camera and points in the scene. To estimate depth for each pixel in the image, we use MegaDepth \cite{DBLP:journals/corr/abs-1804-00607}, a CNN for predicting a dense relative depth map from a single monocular RGB image (see Fig. \ref{fig:An-example-ofMegadepth}). Let $MD(u,v)$ be the relative depth estimated by MegaDepth for the pixel at image coordinates $(u,v)$.

\subsubsection{Pinhole Camera Projection}

The output from MegaDepth is not usable for our purposes as is, and for this reason we create a 3D point cloud from the dense relative depth map. To convert the relative depth values ($MD(u,v)$) estimated by MegaDepth to absolute depth values ($z$), we multiply by an empirically determined scale factor $s$. We transform the depth map of points $(u, v, MD(u,v))$ to a 3D point cloud of points $(x,y,z)$ following the pinhole camera model. 

To simplify the projection between the world coordinate system and the image coordinate system, we assume that the world coordinate system be centered on the camera sensor with the $x$ and $y$ axes aligned with the sides of the sensor plane and the $z$ axis perpendicular to image plane. We assume that the optical center of the camera $(c_x, c_y)$ is at the center of the image and let the world coordinate system origin $(0,0,0)$ be located at the image coordinates origin $(0,0)$. 

Given a relative depth map of points $(u,v, MD(u,v))$ and parameters $s$ and $f$, we create a 3D point cloud of points $(x,y,z)$ by letting 
\begin{align}
    x &= c_x + (u - c_x) \ \frac{z}{f} \label{eq:transx} \\ 
    y &= c_y + (u - c_y) \ \frac{z}{f} \label{eq:transy} \\ 
    z &= s \ MD(u,v) \label{eq:transz}
\end{align}
where $f$ is the focal length of the camera and $s$ is a chosen scaling factor. The next section presents methods for estimating the focal length ($f$) from an image without prior knowledge of the camera used to take the photo.


\subsubsection{Focal Length Estimation}\label{subsec:Focal-length-estimation}

To convert a depth map to a 3D point cloud, we need to know the focal length of the camera. We estimate the focal length from the image using a method based on vanishing point detection from Li et al. \cite{10.1007/978-3-642-17274-8_15}. Li et al. detect lines in the image and use the distribution of the orientations of those lines to estimate vanishing points and then calculate the focal length.

To identify the dominant plane within the crowd area, we use the segmentation mask provided by PSPNet \cite{zhao2016pyramid} to identify points in the 3D reconstruction pertaining to the largest connected crowd component and then apply the RANSAC algorithm \cite{Fischler:1981:RSC:358669.358692}. 

\subsection{Asset Placement}

\begin{figure}
    \centering
    \subfloat[``Unnatural'' alignment ]
	{
	    \frame{\includegraphics[width=0.35\textwidth]{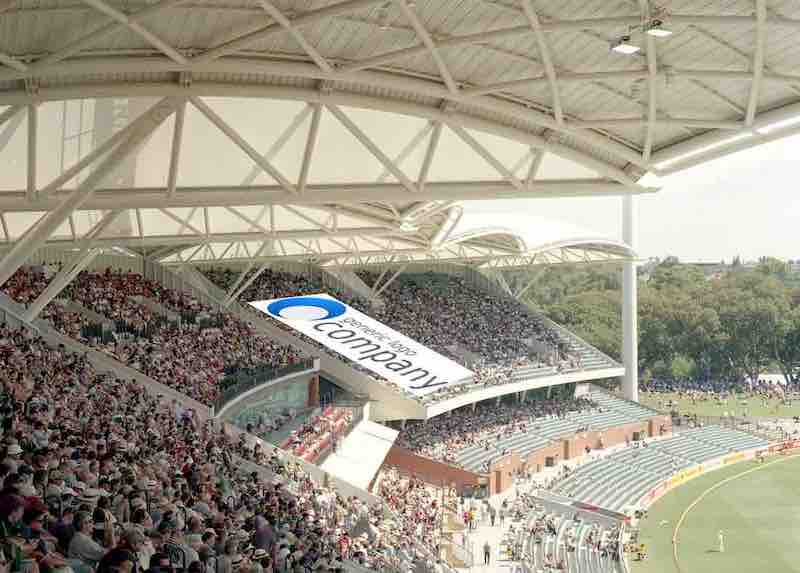}}
    }
    ~
    \subfloat[``Natural'' alignment]
	{
	    \frame{\includegraphics[width=0.35\textwidth]{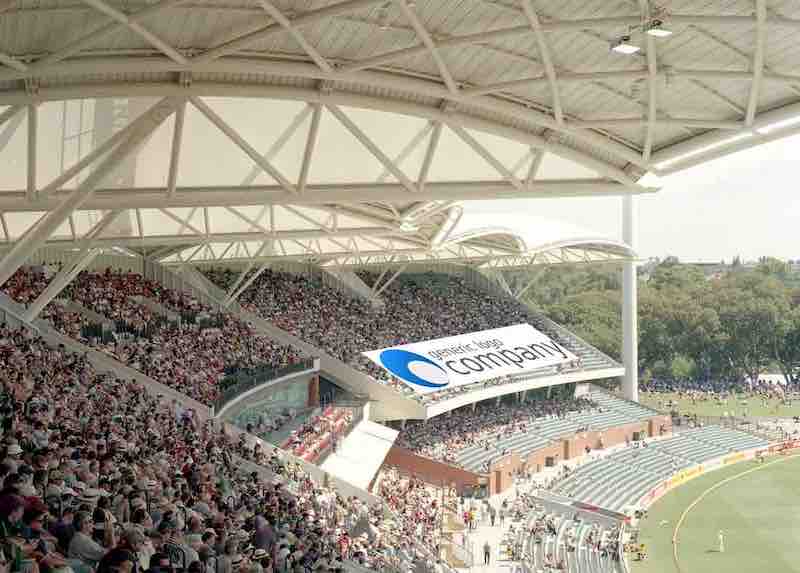}}
    }
    \caption{Perspective correct asset placement with ``unnatural'' and ``natural'' orientation relative to the scene.}
    \label{fig:Correct-asset-placement}
\end{figure}

For the asset to look ``natural", we would like it to appear aligned with the intersection of the crowd plane and the adjacent class (in our case, the ground) plane. In Fig. \ref{fig:Correct-asset-placement}, both the asset placements are technically correct, but only one of them (Fig. \ref{fig:Correct-asset-placement}b) provides a harmonious viewing experience. With this consideration in mind, we aim to align the asset parallel to the dividing line between the crowd and ground planes.

\subsubsection{Orienting the Asset}

We experimented with RANSAC \cite{Fischler:1981:RSC:358669.358692} to fit planes to both the crowd and ground regions with information obtained from the segmentation mask and the depth map. The resulting planes intersect at a 3D line; however, due to noise issues with the depth map, when this line is projected back into the image plane for asset placement, the asset looks ``unnatural''. To uniquely specify the orientation of the asset, we need a vector $\vec{v}$ that is parallel to the surface of the 3D crowd plane to serve as the bottom edge of the rectangular asset. We turned to edge and line detection methods to identify a line to align the asset with. 

\begin{figure}
	\centering
	\subfloat[Binary segmentation]
	{
	    \frame{\includegraphics[width=0.25\textwidth]{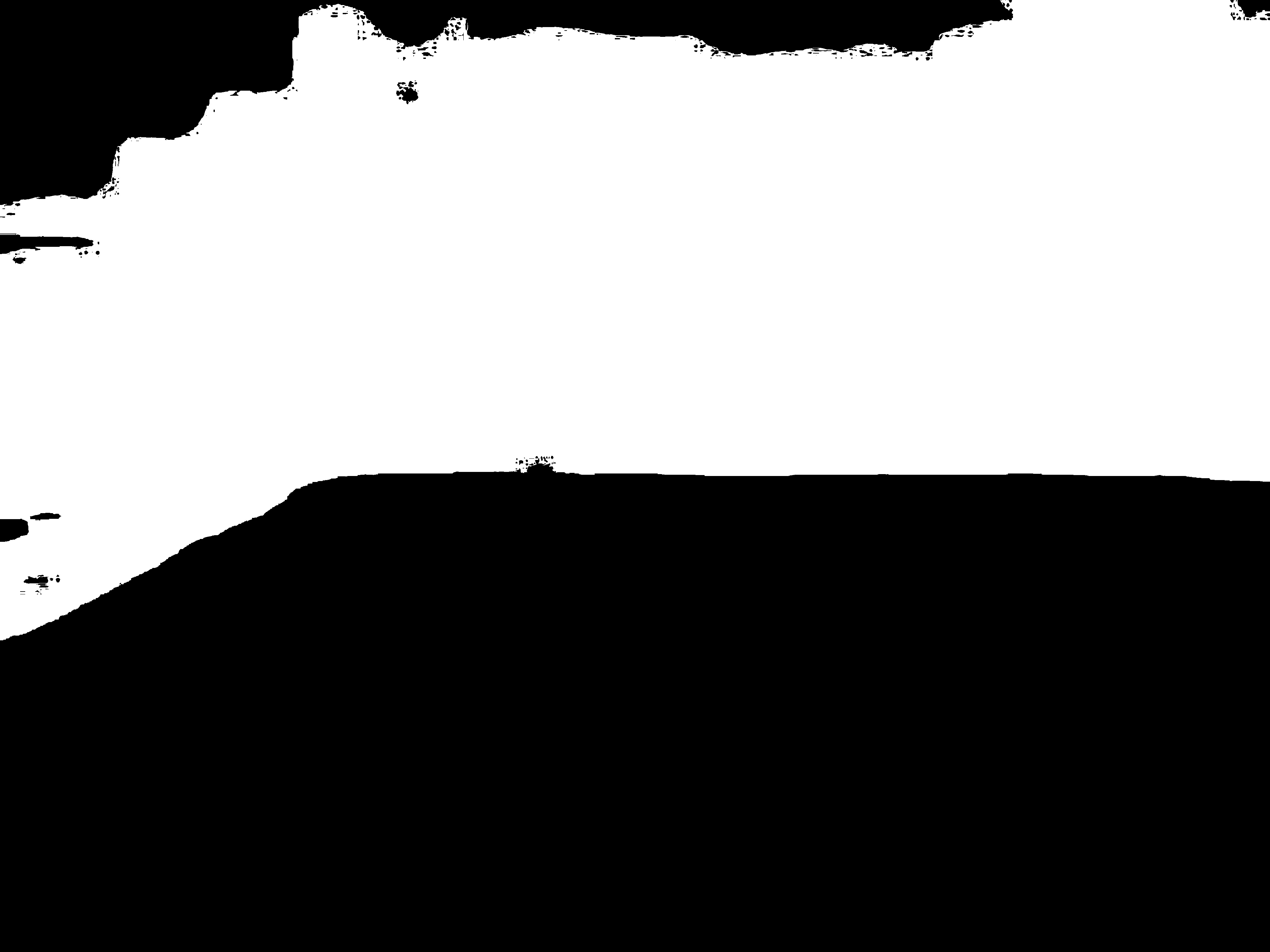}}
	    \label{fig:binary_seg}
	}
	~
	\subfloat[Detected edges with alignment line]
	{
	    \frame{\includegraphics[width=0.25\textwidth]{images/michigan_seg_grandstand_align}}
	    \label{fig:alignment_line}
	} 
	~
	\subfloat[Alignment plane]
	{
	    \includegraphics[width=0.35\textwidth]{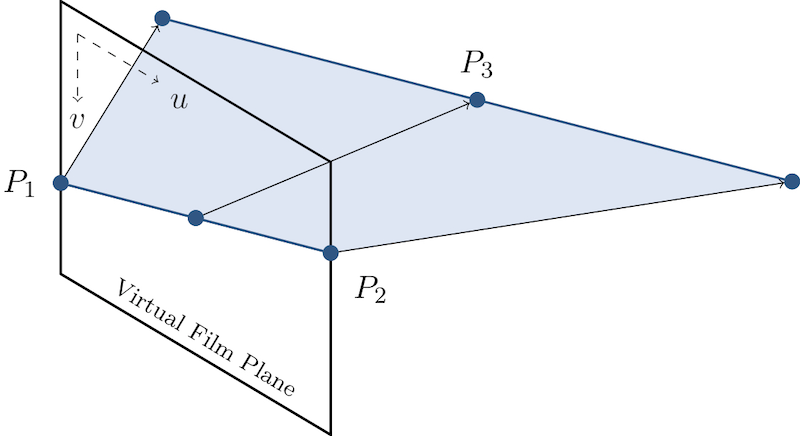}
	    \label{fig:alignment_plane}
	}
	\caption{Illustration of asset placement procedure using Fig. \ref{fig:michigan} as input: (a) The largest crowd component (\emph{white}) identified by segmentation (b) Canny edge detection \cite{Canny:1986:CAE:11274.11275} is applied and the longest line found by Hough line detection \cite{Duda1972} is selected as the alignment line (\emph{red}). (c) The points $P_{1}$ and $P_{2}$ in the image plane are used with a point $P_{3}$ to compute the equation of the alignment plane.}
	\label{fig:Alignment-line-shown}
\end{figure}

We will first find an alignment line in 2D image coordinates, and then transform it into a 3D plane of points that can be inserted into our 3D reconstruction and used to calculate the alignment vector $\vec{v}$. 
To first identify the edges of the crowd region, we apply Canny edge detection \cite{Canny:1986:CAE:11274.11275} to the binary segmentation image (Fig. \ref{fig:binary_seg}) that identifies the largest connected crowd component. having identified the edges of the target crowd area, we now use Hough line detection \cite{Duda1972} to identify straight line segments. We choose the longest line segment to be our alignment line with equation $y = ax + b$ in image coordinates (Fig \ref{fig:alignment_line}).

To align the asset such that is appears parallel to this 2D alignment line in the final image, we must place the asset in 3D parallel to the plane of points which project to the 2D alignment line. We calculate the equation of the 3D alignment plane (Fig. \ref{fig:alignment_plane}) by choosing three arbitrary points ($P_1, P_2, P_3$) on the 2D alignment line in the image plane. We assume that points $P_1$ and $P_2$ in our image have depths of $0$, and point $P_3$ has an arbitrary (non-zero) depth of $z$ in the real world relative to the camera. We can then reconstruct these three sets of image coordinates in 3D following the pinhole camera model as described in Eqns.  \ref{eq:transx}-\ref{eq:transz}. Our 3D reconstruction consists of three points ($P_{1}, P_{2},P_{3}$), which uniquely determine the alignment plane of points which project to the alignment line in the 2D image. 

To make the placed asset appear parallel to the alignment line, we find the intersection of the alignment plane and the dominant crowd plane in the 3D reconstruction. Let vector component of this intersection line be the alignment vector $\vec{v}$. We place the asset in the reconstruction such that it is on the surface of the dominant crowd plane and one of its edges is parallel to the alignment vector ($\vec{v}$).

\subsubsection{Sizing the Asset}
With considerations to maximize the asset size without covering any non-crowd pixels, we fit a convex hull to the inliers of the crowd plane and require all corners of the asset to be inside the hull. The aspect ratio of the overlaid asset in the 3D reconstruction is equivalent to the original aspect ratio of the asset. Finally, we use a homography transformation \cite{Hartley:2003:MVG:861369} to transform the coordinate system of the asset to the coordinate system of the target image, i.e. onto the identified crowd plane. 

\subsection{\label{subsec:Asset-Tracking-Across}Asset Tracking Across Video}

To reduce computation time we augment a seed frame and then employ a tracker to place the asset in all frames. The asset should appear stabilized in a perspective-correct way in subsequent frames, in the face of smooth camera motion and across shot boundaries.

Empirically, we found tracking to be far more accurate if we use four anchor quadrilateral tracks around the asset rather than track the points within the asset boundary. We use a combination of of SIFT \cite{Lowe:2004:DIF:993451.996342}, SURF \cite{Bay:2008:SRF:1370312.1370556} and KAZE \cite{Alcantarilla:2012:KF:2403205.2403222} feature descriptors to obtain key points at each of the corners of the asset's updated location. To track multiple features at once, we make use of the work by Lucas-Kanade on optical flow-based tracking \cite{Lucas:1981:IIR:1623264.1623280}.

\subsubsection{Shot Change Detection}

Since we will be overlaying assets on crowd regions in videos of sporting events, we recognize that these events are commonly filmed using multiple cameras, from various vantage points. The resulting shot changes can occur rapidly and frequently throughout. We detect shot changes per the process in \cite{shotchange}. When a shot change is detected, tracking is suspended but features used for tracking just prior to the shot change are saved. As the video advances, new features are extracted and matched with the stored features to determine if the previously tracked feature points are back in view. If an excessive amount of frames pass and no match is found, the entire system is restarted. We leave the definition of excessive up to the preference of the user. 

\subsubsection{Improving Robustness}

\begin{figure}
	\centering
	\frame{\includegraphics[width=0.5\linewidth]{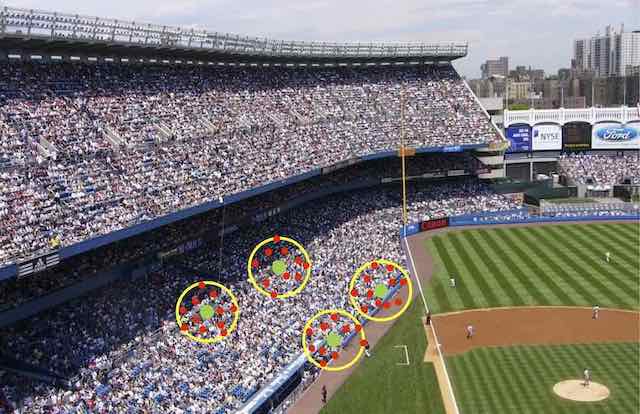}}
	\caption{Features (\emph{red points}) to be tracked are identified within a 50 px radius (\emph{yellow circles}) of the corners (\emph{large green points}) of the quadrilateral.}
	\label{fig:The-red-points}
\end{figure}

Tracking fails when one or more corners of the quadrilateral can no longer be matched with the previous frame. This generally occurs when said corner goes out of frame. To avoid this, we try to predict the position of the corner which is out of frame using: a) the locations of the remaining corners and, b) prior knowledge about the quadrilateral positioning of the corners using the Kalman filter \cite{Kalman}. First, we identify features to track around each corner of the quadrilateral. For example, in Fig. \ref{fig:The-red-points}, we have four groups of features, illustrated by red points. Each group is inside a circle with radius $r = 50px$ centered at each corner of the quadrilateral. For each group of features we calculate the average velocity vector, then set the velocity vector for each corner of the quadrilateral to be
\begin{equation}
    \vec{v_{corner_{i}}} 
    \ = \ 
    \alpha \vec{v_{i}} \ + \ (1 - \alpha) \ \frac{ \vec{v_{1}} +    \vec{v_{2}}  + \vec{v_{3}} + \vec{v_{4}} }{4}
    \label{eq:corner}
\end{equation}
where $\vec{v_{i}}$ is the computed average velocity vector for each
corner $i$, and $\alpha \in (0 , 1)$ is a parameter chosen empirically. We found that $\alpha \geq 0.8$ works best, because the motion of the features in the group centered at a particular corner are more indicative of that corner's movement than the motion of features from the other corner groups.

To calculate the new position of the points which are out of frame,
we compute the Kalman Filter \cite{Kalman} prediction on that corner
and then use $\vec{v_{corner_{i}}}$ from Eqn. \ref{eq:corner} to
update the error estimation.

\section{\label{sec:Conclusion}Conclusion}

\subsection{Results}

\begin{figure}
    \centering
    \subfloat[Input image]
	{
	    \frame{\includegraphics[width=0.3\textwidth]{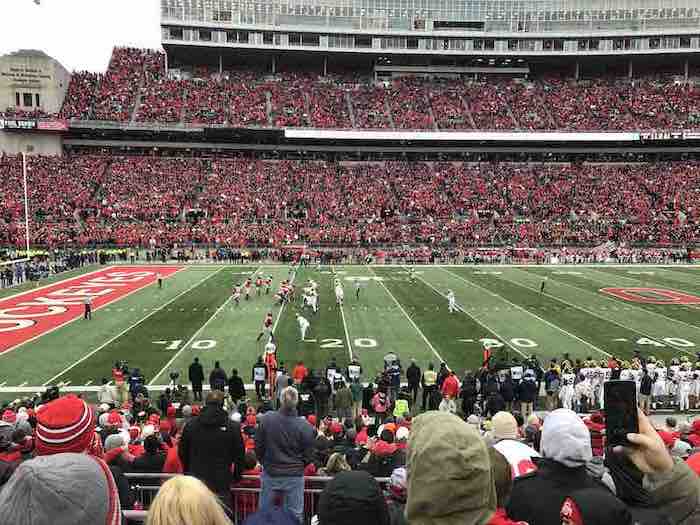}}
	    \label{fig:michigan}
	}
	~
	\subfloat[Segmentation]
	{
	    \frame{\includegraphics[width=0.3\textwidth]{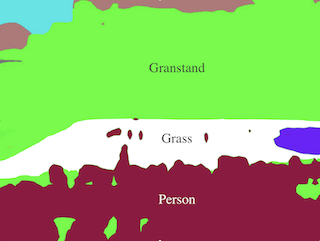}}
	}
	\\
	\subfloat[Largest crowd component]
	{
	    \frame{\includegraphics[width=0.3\textwidth]{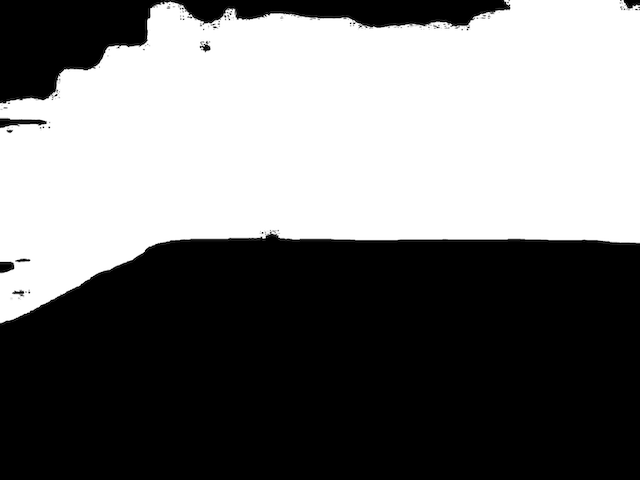}}
	}
	~
	\subfloat[Alignment Line]
	{
	    \frame{\includegraphics[width=0.3\textwidth]{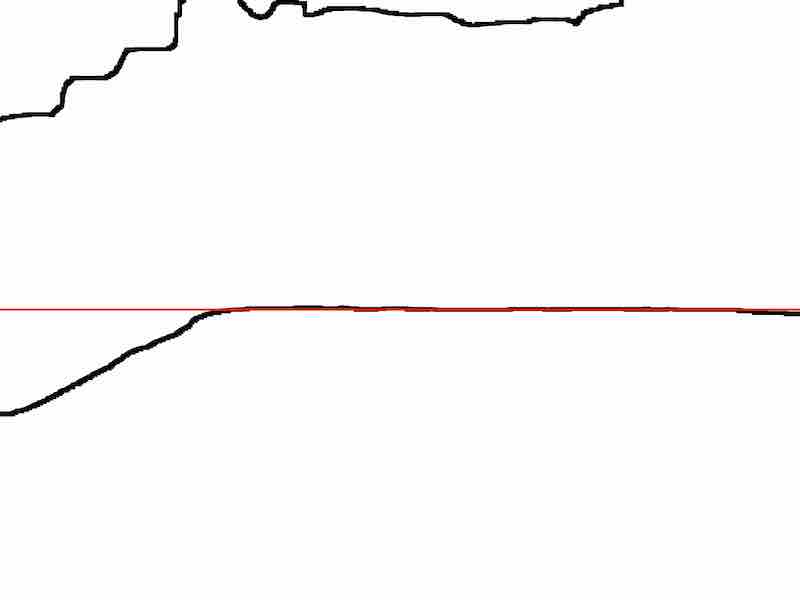}}
	}
	\\
	\subfloat[Depth estimation]
	{
	    \frame{\includegraphics[width=0.3\textwidth]{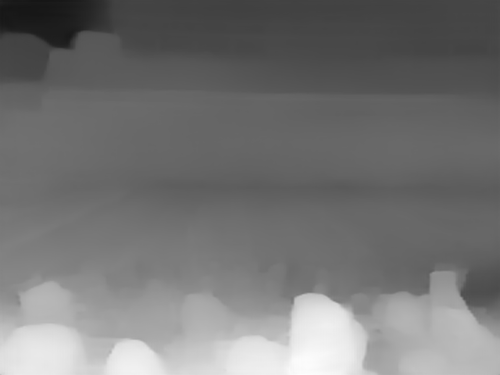}}
	    \label{fig:michigan_depth_map}
	}
	~
	\subfloat[3D reconstruction]
	{
	    \frame{\includegraphics[width=0.3\textwidth, height=0.225\textwidth]{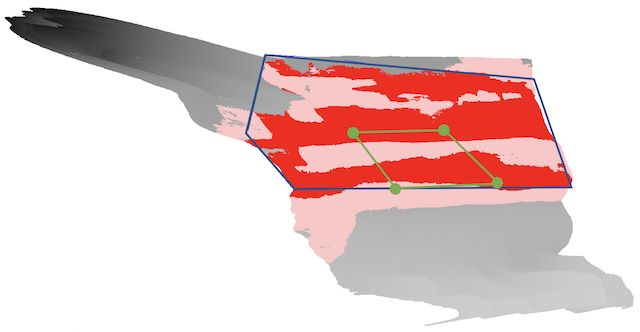}}
	    \label{fig:3d_reconstruction}
	}
	\\
	\subfloat[Crowd plane inliers]
	{
	    \frame{\includegraphics[width=0.3\textwidth]{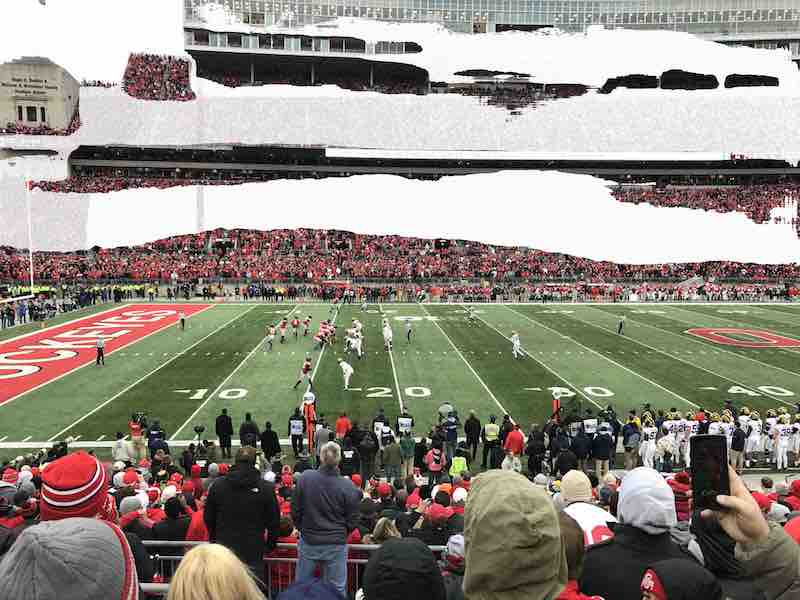}}
	}
	~
	\subfloat[Output image]
	{
	    \frame{\includegraphics[width=0.3\textwidth]{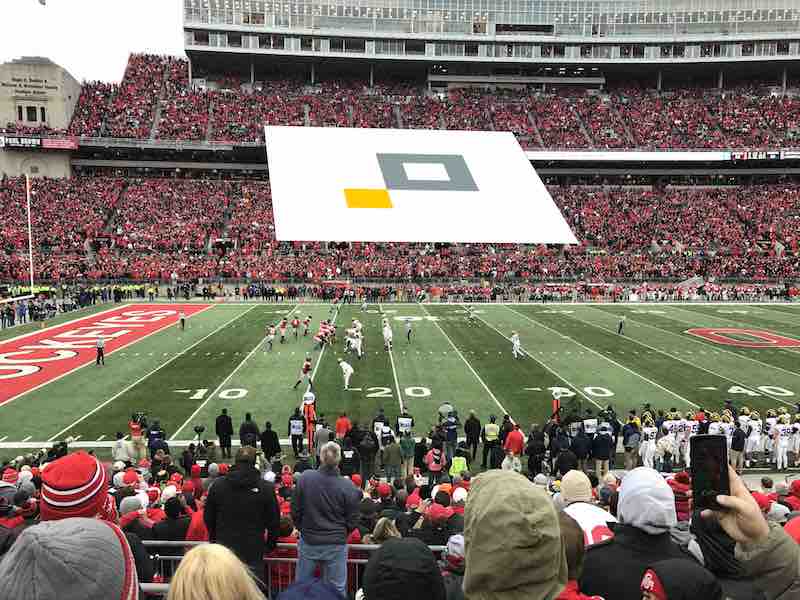}}
	    \label{fig:michigan_output}
	}
    \caption{Example of the pipeline's intermediate outputs running on a single image. The pipeline was run with a depth scaling factor of 1,000,000 and RANSAC tolerance of 10,000. In the 3D point cloud (\ref{fig:3d_reconstruction}) created from the MegaDepth depth map (\ref{fig:michigan_depth_map}) a convex hull (\emph{blue}) is fit to the inliers (\emph{dark red}) of the dominant plane in the segmented crowd region (\emph{light red}). The rectangle (\emph{green}) in the 3D reconstruction (\ref{fig:3d_reconstruction}) corresponds to the final placement of the asset in the output image (\ref{fig:michigan_output}).}
    \label{fig:intermediates}
\end{figure}

Examples of the intermediate results of all of the steps in the pipeline on a single image are illustrated in Fig. \ref{fig:intermediates}. We have in place two pipelines, one for a single image, and one for a video. The pipeline for the single image is the same as the video pipeline but excludes the tracking as well as the SQS (Sec. \ref{subsec:Measuring-segmentation-quality}) calculation.

An example of the system's output for a video is provided online \footnote{\url{https://youtu.be/ugZ-08c6IWY}}. In this demonstration, the asset is also a video, thus augmenting the target crowd region with a video overlay. Video assets appear aesthetically pleasing as long as their framerate matches that of the target video. The entire pipeline was assembled and tested on a 64-bit CPU (Intel Core i7-6800K, 3.40GHz). PSPNet \cite{zhao2016pyramid} was run on a GPU (Nvidia GTX 1080, with 8GB RAM). Table \ref{tab:Timing-per-step} outlines the time taken per step (approximately) to select and augment a seed image of size 1920 x 1080 px from a series of 25 video frames. 

\begin{table}
    \caption{Timing per step in pipeline.}
	\label{tab:Timing-per-step}
    \centering
    \begin{tabular}{|r|l|}
        \hline
        Time (secs) & Process, coding platform/language
    	\\
    	\hline 
    	11.491 & PSPNet on GPU, Python and Tensorflow \cite{tensorflow2015-whitepaper} 
    	\\ 
    	0.374 & SQS on 25 segmented frames (only in video pipeline), Python 
    	\\
    	12.159 & MegaDepth, Python and Torch \cite{torch}
    	\\
    	0.922 & Segment image to select ``crowd`` region, Python
    	\\
    	1.117 & Identify alignment line, Python
    	\\
    	6.627 & Estimate focal length using vanishing points, Octave \cite{eaton:2002}
    	\\
    	8.664 & Create 3D reconstruction and place asset in 3D, C++ 
    	\\
    	0.008 & Display inliers (only for visualization purposes), C++
    	\\
    	0.090 & Warp asset using homography, C++
    	\\ \hline 
    \end{tabular}
\end{table}

\subsection{Limitations}

Many steps need to succeed in order for the system to work. The pipeline performs best on daytime scenes with a single crowd plane. Night-time scenes are challenging for MegaDepth \cite{DBLP:journals/corr/abs-1804-00607} due to the bright overhead lighting often used by sporting arenas. Scenes in which the crowd is covered by an awning are also challenging for MegaDepth as problematic shadows are cast. The focal length estimation method (detailed in Sec. \ref{subsec:Focal-length-estimation}) was designed for images that contain clear orthogonal axes, such as present in a Manhattan world. We observed that the error in estimated focal length was smaller for images taken with small focal lengths. Perpendicular lines will appear more acute when captured with small focal lengths as compared to larger focal lengths which have a wider field of view. 

\subsection{Future Work}

Success of the pipeline relies on accurate depth estimation and image segmentation. Improving these steps may allow for more robust behavior in a wider range of scenarios. Our pipeline utilizes both MegaDepth \cite{DBLP:journals/corr/abs-1804-00607} and PSPNet \cite{zhao2016pyramid} without any major modifications. An alternative technique for depth estimation would be to follow the method of Sturm et al. \cite{10.1007/3-540-61123-1_183}. Producing more reliable and complete camera intrinsic estimates would also allow for the usage of self-localization techniques such as Simultaneous Localization and Mapping (SLAM) \cite{SLAM} to extract depth information. As SLAM is not a learned procedure, its depth estimation will likely be less prone to error in situations where MegaDepth fails to produce adequate results. 

The method of segmentation may be improved upon by overcoming sensitivity to highly non-homogeneous textures. Accurate segmentation of our target texture, crowd, is sensitive to scale and lighting. Exploring a segmentation approach which better accounts for texture information may be beneficial. With the exception of focal length estimation (Sec. \ref{subsec:Focal-length-estimation}) and the SQS score (Sec. \ref{subsec:Measuring-segmentation-quality}), few steps in the pipeline can be assessed quantitatively. Developing a method for quantitatively assessing the quality of an asset placement would make it easier to evaluate the performance of the pipeline in different use cases. 

Our primary method of benchmarking for perspective correctness has been manual inspection of augmented video. This could be improved by finding a qualitative benchmark to compare against. Additionally, conducting a study in which human participants are asked to assess whether an image asset is perspective correct could reveal viewers' tolerance for imperfect asset placement. 


%
%
%
%

\bibliographystyle{splncs04}
\bibliography{accv_bibliography}

\begin{thebibliography}{10}
\providecommand{\url}[1]{\texttt{#1}}
\providecommand{\urlprefix}{URL }
\providecommand{\doi}[1]{https://doi.org/#1}

\bibitem{tensorflow2015-whitepaper}
Abadi, M., Agarwal, A., Barham, P., Brevdo, E., Chen, Z., Citro, C., Corrado,
  G.S., Davis, A., Dean, J., Devin, M., Ghemawat, S., Goodfellow, I., Harp, A.,
  Irving, G., Isard, M., Jia, Y., Jozefowicz, R., Kaiser, L., Kudlur, M.,
  Levenberg, J., Man\'{e}, D., Monga, R., Moore, S., Murray, D., Olah, C.,
  Schuster, M., Shlens, J., Steiner, B., Sutskever, I., Talwar, K., Tucker, P.,
  Vanhoucke, V., Vasudevan, V., Vi\'{e}gas, F., Vinyals, O., Warden, P.,
  Wattenberg, M., Wicke, M., Yu, Y., Zheng, X.: {TensorFlow}: Large-scale
  machine learning on heterogeneous systems (2015),
  \url{https://www.tensorflow.org/}, software available from tensorflow.org

\bibitem{Alcantarilla:2012:KF:2403205.2403222}
Alcantarilla, P.F., Bartoli, A., Davison, A.J.: Kaze features. In: Proc. of the
  12th European Conf. on Computer Vision (ECCV). ECCV'12, vol.~4, pp. 214--227.
  Springer-Verlag, Berlin, Heidelberg (2012).
  \doi{10.1007/978-3-642-33783-3\_16}

\bibitem{Bay:2008:SRF:1370312.1370556}
Bay, H., Ess, A., Tuytelaars, T., Van~Gool, L.: Speeded-up robust features
  (surf). Computer Vision Image Understanding  \textbf{110}(3),  346--359 (jun
  2008). \doi{10.1016/j.cviu.2007.09.014}

\bibitem{Canny:1986:CAE:11274.11275}
Canny, J.: A computational approach to edge detection. IEEE Trans. on Pattern
  Analysis and Machine Intelligence  \textbf{8}(6),  679--698 (1986)

\bibitem{Chang2010VirtualSA}
Chang, C.H., Hsieh, K.Y., Chiang, M.C., Wu, J.L.: Virtual spotlighted
  advertising for tennis videos. J. Visual Commun. and Image Representation
  \textbf{21},  595--612 (2010)

\bibitem{Chang:2008:VVS:1459359.1459500}
Chang, C.H., Hsieh, K.Y., Chung, M.C., Wu, J.L.: Visa: Virtual spotlighted
  advertising. In: Proc. of the 16th ACM Int. Conf. on Multimedia. pp. 837--840
  (2008). \doi{10.1145/1459359.1459500}

\bibitem{torch}
Collobert, R., Kavukcuoglu, K., Farabet, C.: Torch7: A matlab-like environment
  for machine learning. In: BigLearn, NIPS Workshop (2011)

\bibitem{Cordts2016Cityscapes}
Cordts, M., Omran, M., Ramos, S., Rehfeld, T., Enzweiler, M., Benenson, R.,
  Franke, U., Roth, S., Schiele, B.: The cityscapes dataset for semantic urban
  scene understanding. In: Proc. of the IEEE Conf. on Computer Vision and
  Pattern Recognition (CVPR) (2016)

\bibitem{Duda1972}
Duda, R.O., Hart, P.E.: Use of the hough transformation to detect lines and
  curves in pictures. Commun. ACM  \textbf{15}(1),  11--15 (1972).
  \doi{10.1145/361237.361242}

\bibitem{SLAM}
Durrant-whyte, H., Bailey, T.: Simultaneous localization and mapping: Part i.
  IEEE Robotics Automation Magazine  \textbf{13},  99 -- 110 (2006).
  \doi{10.1109/MRA.2006.1638022}

\bibitem{eaton:2002}
Eaton, J.W.: GNU Octave Manual. Network Theory Limited (2002)

\bibitem{pascal-voc-2012}
Everingham, M., Van~Gool, L., Williams, C.K.I., Winn, J., Zisserman, A.: The
  {PASCAL} {V}isual {O}bject {C}lasses {C}hallenge 2012 {(VOC2012)} {R}esults
  (2012), \url{http://host.robots.ox.ac.uk/pascal/VOC/voc2012/}

\bibitem{Fischler:1981:RSC:358669.358692}
Fischler, M.A., Bolles, R.C.: Random sample consensus: A paradigm for model
  fitting with applications to image analysis and automated cartography.
  Commun. ACM  \textbf{24}(6),  381--395 (1981). \doi{10.1145/358669.358692}

\bibitem{10.1007/978-3-540-73417-8_52}
Han, J., de~With, P.H.N.: 3-d camera modeling and its applications in sports
  broadcast video analysis. In: Multimedia Content Analysis and Mining. pp.
  434--443. Springer Berlin Heidelberg, Berlin, Heidelberg (2007)

\bibitem{Hartley:2003:MVG:861369}
Hartley, R., Zisserman, A.: Multiple View Geometry in Computer Vision.
  Cambridge University Press, New York, NY, USA, 2 edn. (2003)

\bibitem{Kalman}
Kalman, R.: A new approach to linear filtering and prediction problems. J. of
  Basic Engineering (ASME)  \textbf{82D},  35--45 (01 1960)

\bibitem{10.1007/978-3-642-17274-8_15}
Li, B., Peng, K., Ying, X., Zha, H.: Simultaneous vanishing point detection and
  camera calibration from single images. In: Advances in Visual Computing. pp.
  151--160. Springer Berlin Heidelberg (2010)

\bibitem{Li:2005:RTA:1101149.1101221}
Li, Y., Wan, K.W., Yan, X., Xu, C.: Real time advertisement insertion in
  baseball video based on advertisement effect. In: Proc. of the 13th Annual
  ACM Int. Conf. on Multimedia. pp. 343--346 (2005).
  \doi{10.1145/1101149.1101221}

\bibitem{DBLP:journals/corr/abs-1804-00607}
Li, Z., Snavely, N.: Megadepth: Learning single-view depth prediction from
  internet photos. In: Proc. of the IEEE Conf. on Computer Vision and Pattern
  Recognition (CVPR) (2018)

\bibitem{Liu2009AdvertiseG}
Liu, H., Qiu, X., Huang, Q., Jiang, S., Xu, C.: Advertise gently - in-image
  advertising with low intrusiveness. 16th IEEE Int. Conf. on Image Process.
  (ICIP) pp. 3105--3108 (2009)

\bibitem{Lowe:2004:DIF:993451.996342}
Lowe, D.G.: Distinctive image features from scale-invariant keypoints. Int. J.
  Computer Vision  \textbf{60}(2),  91--110 (2004).
  \doi{10.1023/B:VISI.0000029664.99615.94}

\bibitem{Lucas:1981:IIR:1623264.1623280}
Lucas, B.D., Kanade, T.: An iterative image registration technique with an
  application to stereo vision. In: Proc. of the 7th Int. Joint Conf. on
  Artificial Intelligence. IJCAI'81, vol.~2, pp. 674--679. Morgan Kaufmann
  Publishers Inc., San Francisco, CA, USA (1981)

\bibitem{10.1007/978-1-4471-0859-7_6}
Medioni, G., Guy, G., Rom, H., Fran{\c{c}}ois, A.: Real-time billboard
  substitution in a video stream. In: Multimedia Commun. pp. 71--84. Springer
  London (1999)

\bibitem{Mei:2010:ATC:2736420.2736646}
Mei, T., Guo, J., Hua, X.S., Liu, F.: Adon: Toward contextual overlay in-video
  advertising. Multimedia Syst.  \textbf{16}(4-5),  335--344 (2010)

\bibitem{Mei:2008:CIA:1459359.1459418}
Mei, T., Hua, X.S., Li, S.: Contextual in-image advertising. In: Proc. of the
  16th ACM Int. Conf. on Multimedia. pp. 439--448. ACM (2008).
  \doi{10.1145/1459359.1459418}

\bibitem{ILSVRC15}
Russakovsky, O., Deng, J., Su, H., Krause, J., Satheesh, S., Ma, S., Huang, Z.,
  Karpathy, A., Khosla, A., Bernstein, M., Berg, A.C., Fei-Fei, L.: {ImageNet
  Large Scale Visual Recognition Challenge}. Int J. of Computer Vision (IJCV)
  \textbf{115}(3),  211--252 (2015)

\bibitem{10.1007/3-540-61123-1_183}
Sturm, P., Triggs, B.: A factorization based algorithm for multi-image
  projective structure and motion. In: Proc. of the 4th European Conf. on
  Computer Vision (ECCV). pp. 709--720. ECCV '96, Springer Berlin Heidelberg,
  Berlin, Heidelberg (1996)

\bibitem{Wan2006AutomaticCP}
Wan, K.W., Xu, C.: Automatic content placement in sports highlights. 2006 IEEE
  Int. Conf. on Multimedia and Expo pp. 1893--1896 (2006)

\bibitem{10.1007/978-3-540-30542-2_33}
Xu, C., Wan, K.W., Bui, S.H., Tian, Q.: Implanting virtual advertisement into
  broadcast soccer video. In: Adv. in Multimedia Inf. Process. - PCM 2004. pp.
  264--271. Springer Berlin Heidelberg (2005)

\bibitem{shotchange}
Yildrim, Y.: Shotdetection.
  \url{https://github.com/yasinyildirim/ShotDetection} (2015)

\bibitem{zhao2016pyramid}
Zhao, H., Shi, J., Qi, X., Wang, X., Jia, J.: Pyramid scene parsing network.
  In: Proc. of the IEEE Conf. on Computer Vision and Pattern Recognition (CVPR)
  (2017)

\bibitem{zhou2017scene}
Zhou, B., Zhao, H., Puig, X., Fidler, S., Barriuso, A., Torralba, A.: Scene
  parsing through ade20k dataset. In: Proc. of the IEEE Conf. on Computer
  Vision and Pattern Recognition (CVPR) (2017)

\bibitem{zhou2016semantic}
Zhou, B., Zhao, H., Puig, X., Xiao, T., Fidler, S., Barriuso, A., Torralba, A.:
  Semantic understanding of scenes through the ade20k dataset. Int. J. of
  Computer Vision  (2018). \doi{10.1007/s11263-018-1140-0}

\end{thebibliography}

\end{document}